\documentclass{article}
\usepackage{spconf,amsmath,graphicx}
\usepackage[font=small,labelfont=bf]{caption}
\usepackage[table]{xcolor}
\definecolor{Gray}{rgb}{0.88,1,1}

\newcolumntype{a}{>{\columncolor{Gray}}c}
\title{TWO HEADED DRAGONS: MULTIMODAL FUSION AND CROSS MODAL TRANSACTIONS}
\name{Rupak Bose$^*$, Shivam Pande, Biplab Banerjee\thanks{*Corresponding author: boserupak1@gmail.com}}
\address{Centre of Studies in Resources Engineering, Indian Institute of Technology Bombay, India}

\begin{document}

\maketitle

\begin{abstract}
As the field of remote sensing is evolving, we witness the accumulation of information from several modalities, such as multispectral (MS), hyperspectral (HSI), LiDAR etc. Each of these modalities possess its own distinct characteristics and when combined synergistically, perform very well in the recognition and classification tasks. However, fusing multiple modalities in remote sensing is cumbersome due to highly disparate domains. Furthermore, the existing methods do not facilitate cross-modal interactions. To this end, we propose a novel transformer based fusion method for HSI and LiDAR modalities. The model is composed of stacked auto encoders that harness the cross key-value pairs for HSI and LiDAR, thus establishing a communication between the two modalities, while simultaneously using the CNNs to extract the spectral and spatial information from HSI and LiDAR. We test our model on Houston (Data Fusion Contest – 2013) and MUUFL Gulfport datasets and achieve competitive results.
\end{abstract}
\begin{keywords}
Hyperspectral, LiDAR, domain mapping, multimodal fusion, cross-modal inferences
\end{keywords}
\section{Introduction}
\label{sec:intro}

\begin{figure}
    \centering
    \includegraphics[width=8cm]{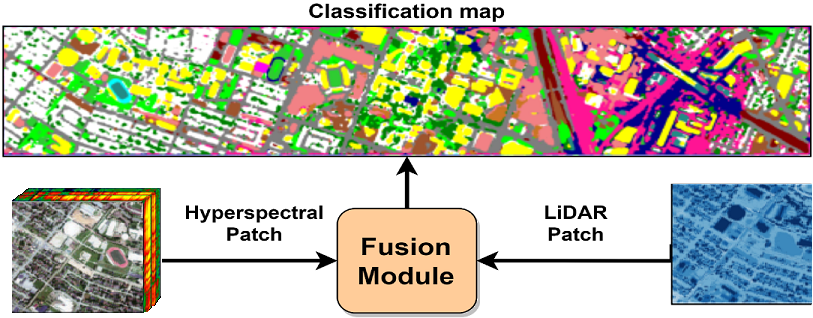}
    \caption{Traditional HSI-LiDAR fusion module. The HSI and LiDAR features are concatenated and sent for classification. }
    \label{fig:concept}
\end{figure}

The existing single modal deep learning classifiers perform well on classification tasks. Having said that, they still have a huge scope of improvements if presented with a complementary dataset such as a corresponding LiDAR digital surface model (DSM) map of the area. With multiple modalities, the classification accuracies of the single modal classifiers are boosted significantly. For instance, the case where the reflectance information is not sufficient to identify areas with similar reflectances, such as water and shadow. In those cases, additional information (such as depth from LiDAR images), can help in classification \cite{mohla2020fusatnet}. Thus, having a fusion based model can learn the mapping of hyperspectral and LiDAR attributes to a common space, where it can infer a combined knowledge for better classification. The general approaches mainly revolve around two stream CNN structures which at some point exploit data fusion parameters for achieving high accuracies (refer Fig. \ref{fig:concept}).
The use of optimal data encapsulation for feature reconstruction is demonstrated to improve accuracies beyond the traditional approaches \cite{hong2020deep}. This method establishes the elegance of reconstruction using concatenated deep features that can hugely improve classification performance. Approach \cite{chen2017deep} being a simple yet innovative architecture performs proficiently and paved way towards concatenation based classification. \cite{mohla2020fusatnet} further improved the fusion performance by introducing visual attention in HSI-LiDAR fusion. The novel attention method helps in highlighting important data features that would be otherwise lost. It was to our knowledge, the first attempt to introduce \textit{cross-modal} attention for initiating interaction between different modalities. The above papers have explored the different mixtures of machine learning/deep learning approaches and have proved that the deep fusion of features works best in a two stream deployment. Other methods like \cite{li2018hyperspectral, feng2019multisource} which have already been tried and tested in the field of data fusion, prove their merits in image classification with little cross-modal interaction.

Since, the previous approaches perform fusion either at the beginning or at the intermediate level, their models are unable to effectively capture the representations in the fused space. Hence, we go for cross-modal interaction at different levels to fuse high and low level features, that are important for classification. To this end, we draw the idea from the \textit{transformer} architecture \cite{vaswani2017attention} which played a key role in using cross-attention as well as self-attention for long sequence NLP tasks. Having seen the \textit{transformer} define the state of the art for various cross-domain tasks like machine translation and sequence to sequence modelling, we attempt to solve a classic \textbf{cross-domain fusion} problem. Our contributions in this paper would be:

\begin{itemize}
    \item We redefine \textit{queries} with \textit{cross key-value} pairs for multimodal data fusion of HSI and LiDAR images as images of different domains belong in the same geographical location. 
    \item We utilize CNNs to capture spatial data while using attention modes to capture the spectral essence and to achieve a true spatio-spectral fusion module.
    \item We introduce a modular design for an attempt at maximum cross modal interaction with spectro-spatial feature selection based on attention with dual modalities.
\end{itemize}

\section{METHODOLOGY}
\label{sec:method}

We propose a stacked encoder-decoder architecture where attention based cross modal information is preserved. The filter block generates a representative spectral map based on spatial features, such as \textit{queries}, \textit{keys} and \textit{values} for their respective inputs. The spectral attention decided on the basis of \textit{queries} and \textit{keys} of same modality is captured in the self attention blocks, which scores closely related pixels higher, thereby giving attention in the spectral zone. The cross-attention decoder takes the attention scores and decodes the corresponding \textit{cross key-value} pairs by highlighting the important information based on the input attention scores. The `CrossOut' (section \ref{sec:decoder}) ensures that the input data is preserved through the decoding. The architecture being modular in design, aides in easy stacking of the blocks. As \textit{key-value} pairs are cross-modal, there we exploit the cross-modal interactions.
\subsection{Encoder block}
\begin{figure}
    \centering
    \includegraphics[width=8.5cm]{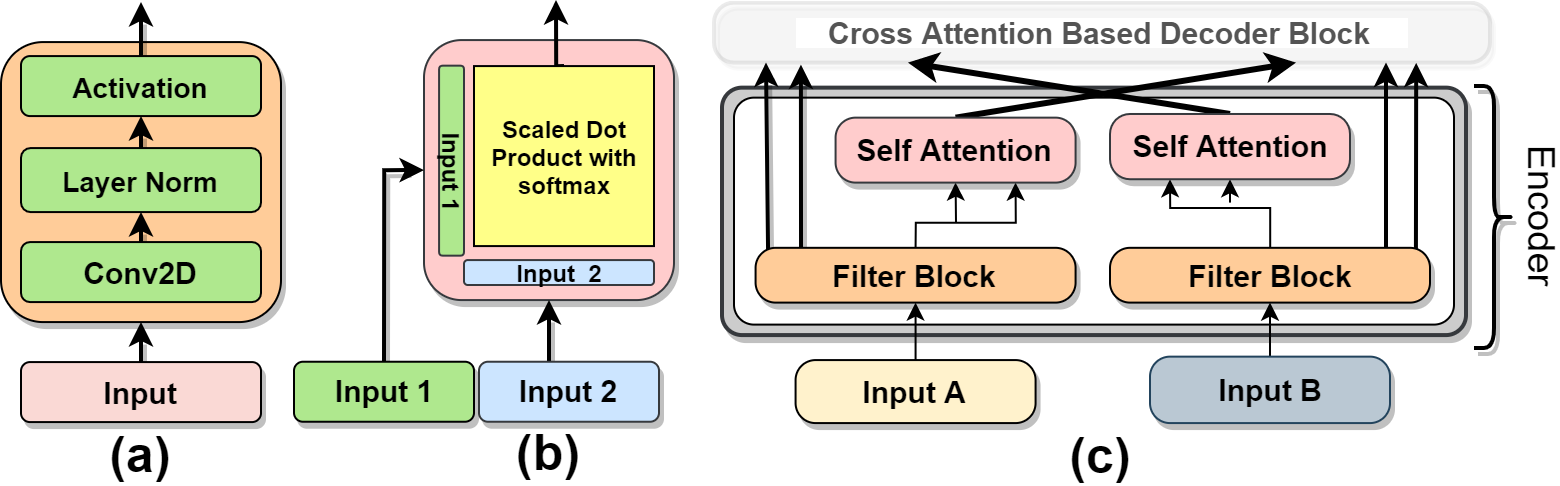}
    \caption{Encoder block components. (a) Filter block, (b) Self attention module, (c) Encoder block. The inputs for the encoder block are the originals modalities (when placed in the beginning) or the outputs of the previous decoders (for all subsequent encoders). }
    \label{fig:encoder}
\end{figure}
Encoder block [Fig. \ref{fig:encoder} (c)] consists of 2 components, namely, Filter Block and Self Attention Module. This block generates the \textit{queries} and corresponding \textit{key-value} pairs to generate attention score based weights (inspired from \textit{transformers} \cite{vaswani2017attention}).

\noindent\textbf{Filter Block}: It is a {Conv2D → Layer Normalization → Activation} implementation [Fig. \ref{fig:encoder} (a)], where convolution is done with padding and kernel size is set to 3 with number of filters as ‘$d$’. The filter block being a convolutional block captures the spatial data for both the modalities. The outputs of the filter blocks are their corresponding \textit{queries} with \textit{key-value} pairs. The convolutional filter block ensures that the generated \textit{key-value} pairs are in the same geolocation but have individual channel signatures. The \textit{query-key-value} triplets are used for computing self-attention and cross-attention features.

\noindent\textbf{Self Attention Module}: The Self Attention block [Fig. \ref{fig:encoder} (b)] is the attention mechanism inspired from the multiplicative attention based on \textit{luong attention} \cite{luong2015effective}. The scaled dot product attention is defined as $Q\odot K^T/\sqrt{d}$ (‘$\odot$’ stands for the vector dot product). It takes \textit{queries} ($Q$) and \textit{keys} ($K$) as inputs, computes the scaled dot product and performs \textit{softmax} on it. The scaled dot product based attention maps similarity in the spectral space and the softmax layer pushes the row-wise attention scores to (0, 1). The scaling helps is managing the dot products when they tend to get high values \cite{vaswani2017attention}.

\subsection{Decoder block}
\label{sec:decoder}
\begin{figure}
    \centering
    \includegraphics[width=8.5cm]{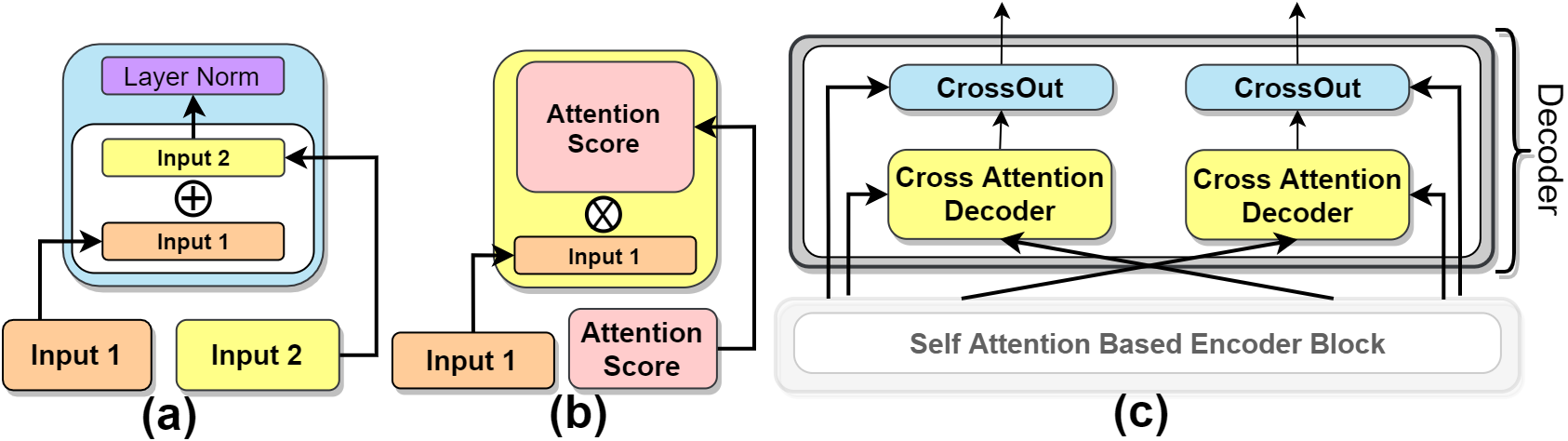}
    \caption{Decoder block components. (a) CrossOut block, (b) Cross-attention module, (c) Decoder block. Here, the input comes from the attention layers and filter blocks of the encoder.}
    \label{fig:decoder}
\end{figure}
The decoder [Fig. \ref{fig:decoder} (c)] in its entirety, decodes the \textit{cross key-value} pairs effectively and select attention based features using the cross decoder module. The CrossOut module ensures that the input modal data is retained along with highlighted cross-modal features to optimally capture data. 

\noindent\textbf{Cross-Attention Decoder}: It [Fig. \ref{fig:decoder} (b)] implements the cross-decoding based on prior \textit{key-value} pairs. It is an $A \odot V$ implementation. $A$ is the cross-domain self-attention score and $V$ is the value generated from the respective filter block.

\noindent\textbf{CrossOut}: It is a residual block implementation that follows \{Input1 + Input2\} followed by layer normalization \cite{ba2016layer} [Fig. \ref{fig:decoder} (a)]. This ensures that the abstractions includes features from both cross and individual domains. 

\subsection{The full stack: Two headed dragon} 
\begin{figure}
    \centering
    \includegraphics[width=8cm]{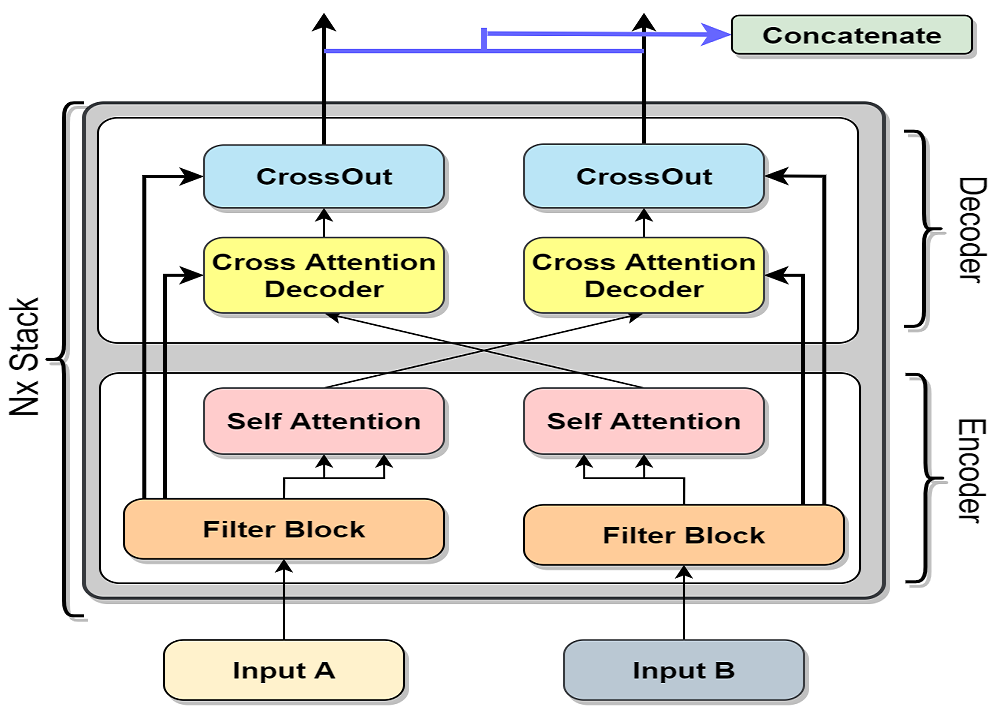}
    \caption{Schematic of full stack with both encoder and decoder.}
    \label{fig:full_stack}
\end{figure}
The full stack [Fig. \ref{fig:full_stack}] is the stacking of the Encoder Block and the Decoder Block in a way that cross-attention is achieved. The number of stacks ($N_X$) can be varied as required. Also, this being a modular build, the number of filters for individual filter blocks can also be user defined. The stacking also aids in multiple cross interaction with self signature preservation. The two step attention process selectively filters the relevant and important information to be preserved for the next step. There are two outputs which are complementary yet have interacted using cross-attention, hence the name “Two headed dragon”.

\subsection{Proposed architecture}
\begin{figure}
    \centering
    \includegraphics[width=8cm]{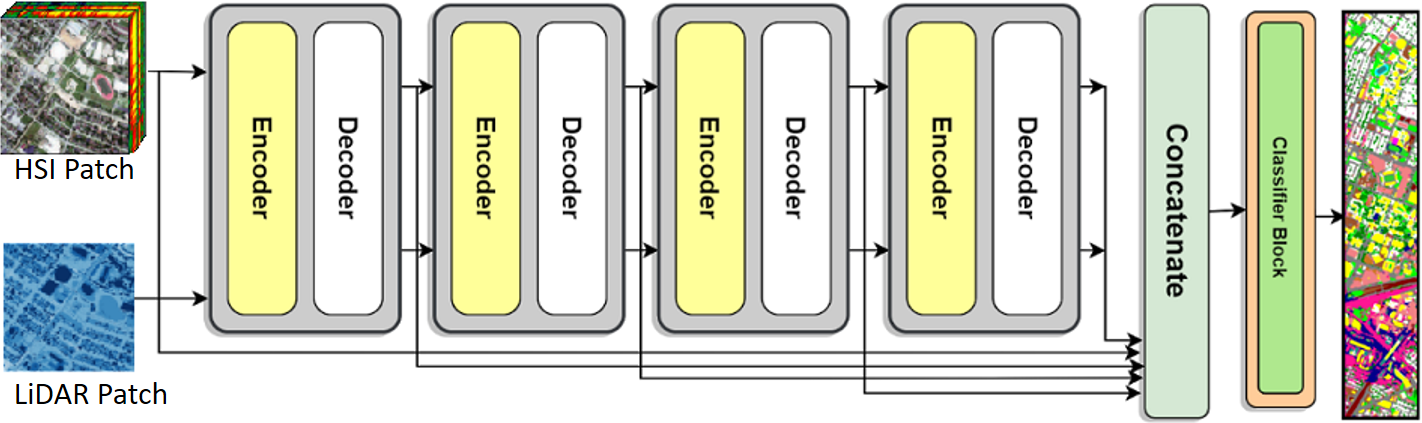}
    \caption{Schematic of complete fusion based classification model.}
    \label{fig:model}
\end{figure}
We use a 4 Stacked Encoder - Decoder architecture [Fig. \ref{fig:model}], where the output from all the stacks is concatenated, which represents a global common space representation for the HSI and LiDAR patches. This concatenated feature passes through a classification block for classification. The classification block is a {Global Average Pooling → Dropout → Softmax (number of classes)} implementation. The entire model is trained on categorical crossentropy loss, given as:
\begin{equation}
\mathcal{L}_{CE} = -\sum_{i=1}^{N} y^i\log{\hat{y}^i}
\label{equation:ce}
\end{equation}
where, $N$ is the number of training samples, $y^i$ and $\hat{y}^i$ are respectively groundtruth class and prediction for the $i^{th}$ sample.

\begin{table}[ht]
\centering{\scriptsize
 \caption{\label{tab:H13_perf} Accuracy analysis of 2013 Houston dataset (in \%). ‘H’ represents only HSI and ‘H+L’ represents fused HSI and LiDAR.}
\begin{tabular}{|p{0.8cm}| p{0.9cm} |p{1.3cm} |p{1.1cm} |p{0.7cm} |p{0.7cm}|}
 \hline
Class & EncDec \cite{hong2020deep} & Two Branch CNN \cite{xu2017multisource} & Deep Fusion \cite{chen2017deep} & FusAt Net \cite{mohla2020fusatnet} & Ours \\
 \hline
    1     & 82.81 & 82.62 & \textbf{82.87} & 82.53 & 82.05 \\
    2     & 83.65 & 84.49 & 84.87 & \textbf{85.15} & 84.68 \\
    3     & 97.62 & 81.58 & 96.44 & 99.21 & \textbf{99.80} \\
    4     & 92.52 & \textbf{99.43} & 92.14 & 92.80  & 97.63 \\
    5     & \textbf{100}   & 99.53 & 99.43 & \textbf{100} & \textbf{100} \\
    6     & 87.41 & 81.12 & 93.01 & \textbf{100}   & 96.50 \\
    7     & 86.01 & 78.26 & 87.31 & \textbf{94.87} & 88.71 \\
    8     & 82.91 & 87.94 & 78.82 & 83.48 & \textbf{89.55} \\
    9     & 80.26 & 83.66 & 88.01 & 85.93 & \textbf{91.60} \\
    10    & 66.89 & 63.22 & \textbf{80.98} & 68.05 & 68.92 \\
    11    & 89.28 & 88.71 & 92.41 & 90.32 & \textbf{96.77} \\
    12    & 96.54 & 88.38 & 84.34 & 92.51 & \textbf{97.02} \\
    13    & 72.63 & 91.93 & 85.96 & \textbf{92.63} & 88.77 \\
    14    & 96.55 & 78.14 & 89.88 & 99.60  & \textbf{100} \\
    15    & 97.25 & 78.44 & \textbf{97.89} & 96.41 & 95.77 \\ \hline
    OA    & 86.90  & 85.14 & 88.02 & 88.93 & \textbf{90.64} \\
    Only H & 83.82 & 79.32 & 84.07 & 85.15 & \textbf{85.98} \\ \hline
\end{tabular}}
\end{table}

\begin{table}[ht]
\centering{\scriptsize
 \caption{\label{tab:MU_perf} Accuracy analysis of MUUFL dataset (in \%). ‘H’ represents only HSI and ‘H+L’ represents fused HSI and LiDAR.}
\begin{tabular}{|p{0.8cm}| p{0.9cm} |p{1.3cm} |p{1.1cm} |p{0.7cm} |p{0.7cm}|}
 \hline
Class & EncDec \cite{hong2020deep} & Two Branch CNN \cite{xu2017multisource} & Deep Fusion \cite{chen2017deep} & FusAt Net \cite{mohla2020fusatnet} & Ours \\
 \hline
    1     & 89.84 & 88.39 & 87.96 & 91.61 & \textbf{93.80} \\
    2     & 83.74 & 75.95 & 79.88 & 72.01 & \textbf{85.56} \\
    3     & 69.68 & 75.92 & 77.38 & 81.42 & \textbf{82.87} \\
    4     & 93.22 & 93.74 & 95.02 & \textbf{97.16} & 96.23 \\
    5     & 86.50  & 51.06 & 89.45 & 89.72 & \textbf{91.76} \\
    6     & \textbf{100}   & 98.91 & \textbf{100}   & 99.45 & 99.73 \\
    7     & 92.45 & 93.25 & 92.17 & \textbf{96.20}  & 90.62 \\
    8     & 95.77 & 94.38 & 92.83 & 94.58 & \textbf{97.43} \\
    9     & 77.12 & \textbf{82.72} & 67.63 & 80.47 & 78.29 \\
    10    & 91.57 & 91.57 & 91.57 & 90.36 & \textbf{95.18} \\
    11    & 96.45 & 97.04 & \textbf{99.41} & \textbf{99.41} & \textbf{99.41} \\ \hline
    OA    & 87.03 & 82.16 & 86.94 & 89.02 & \textbf{91.64} \\
    Only H & 86.84 & 81.57 & 85.16 & 88.03 & \textbf{90.02} \\ \hline
\end{tabular}}
\end{table}

\begin{table}[ht]
\centering{\scriptsize
 \caption{\label{tab:abl1} Ablation study for change in stack number.}
\begin{tabular}{|p{1.6cm}| p{0.5cm} |p{0.5cm} |p{0.5cm} |p{0.5cm}|}
 \hline
 Stack ($N_x$) & 1     & 2     & 3     & 4 \\
 \hline
    Houston & 87.63 & 89.49 & 89.73 & \textbf{90.64} \\
    MUUFL & 85.49 & 88.18 & 90.06 & \textbf{91.64} \\\hline
\end{tabular}}
\end{table}

\begin{table}[ht]
\centering{\scriptsize
 \caption{\label{tab:abl2} Ablation study for change in embedding size.}
\begin{tabular}{|p{1.6cm}| p{0.5cm} |p{0.5cm} |p{0.5cm} |p{0.5cm} |p{0.5cm}|}
 \hline
 Embed Size & 64    & 96    & 128   & 160   & 192 \\
 \hline
    Houston & 86.13 & 87.61 & \textbf{90.64} & 90.36 & 90.16 \\
    MUUFL & 88.80  & 89.19 & \textbf{91.64} & 90.62 & 90.25 \\ \hline
\end{tabular}}
\end{table}

\section{PERFORMANCE EVALUATION}

\begin{figure*}
    \centering
    \includegraphics[width=17.5cm]{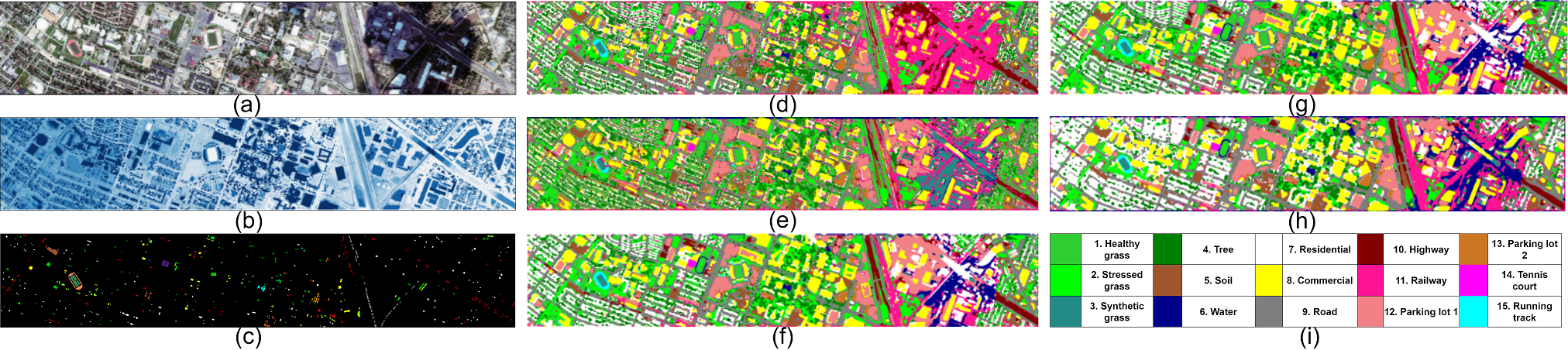}
    \caption{Dataset. (a) RGB FCC, (b) LiDAR, (c) Ground Truth. Classification maps. (d) Deep Encoder Decoder, (e) Two Branch CNN, (f) Deep Fusion, (g) FusAtNet, (h) Ours, (i) Color Scheme}
    \label{fig:houston}
\end{figure*}

\subsection{Training protocols} 

The input image patches are of size 11$\times$11$\times$channels. We compare our model's performance with other state of the art HSI-LiDAR fusion models from \cite{hong2020deep} (encoder decoder based model, shortened as EncDec), two branch CNN \cite{xu2017multisource}, deep fusion \cite{chen2017deep} and FusAtNet \cite{mohla2020fusatnet}. We use Adam optimizer \cite{kingma2014adam} for training all the models, where learning rate is set to 5$\times$10$^{-6}$ for a total of 500 epochs with glorot weight initialization \cite{glorot2010understanding}. 

\subsection{Houston 2013 dataset}

The Houston 2013 dataset, one of the largest available HSI+LiDAR datasets, was introduced in GRSS Data Fusion Contest 2013. The HSI image consists of a total of 144 hyperspectral channels (0.38 µm to 1.05 µm) containing [349$\times$1905] pixels per channel. The corresponding LiDAR image is a single channel image with [349$\times$1905] pixels. The total number of ground truth points are 15029 from 15 different classes, of which 2832 samples are in the training set and the remaining in the test set \cite{mohla2020fusatnet}.

As seen from Table \ref{tab:H13_perf}, DeepFusion does exceptionally well for the \textit{Highway} class (10) which is a bit problematic in other architectures including our proposed method. FusAtNet performs great for the classes 2, 5, 6, 7 and 13, major boost being in the \textit{Residential} class. It captures very well the spatial relationships of residential areas which can be seen through the classification map [Fig. \ref{fig:houston} (c)]. Two Branch CNN performs well in class 4 by a good margin. However, EncDec even though has a relatively lower accuracy, is the only one that gives a steady increase in accuracy throughout its epochs given the learning rate. Our proposed model outperforms the other models in the classes 3, 5, 8, 9, 11, 12 and 14, with major accuracy boosts for \textit{Trees}, \textit{Commercial}, \textit{Road}, \textit{Railway} and \textit{Parking} classes. The major pitfalls of our proposed model are in the classes of \textit{Highway}, \textit{Water} and \textit{Residential}. It can be seen from our classification map that \textit{Trees}, \textit{Railways} and \textit{Roads} have a definite form, however lack in the \textit{Residential} and \textit{Water} class depiction. 

\begin{figure}
    \centering
    \includegraphics[width=8cm]{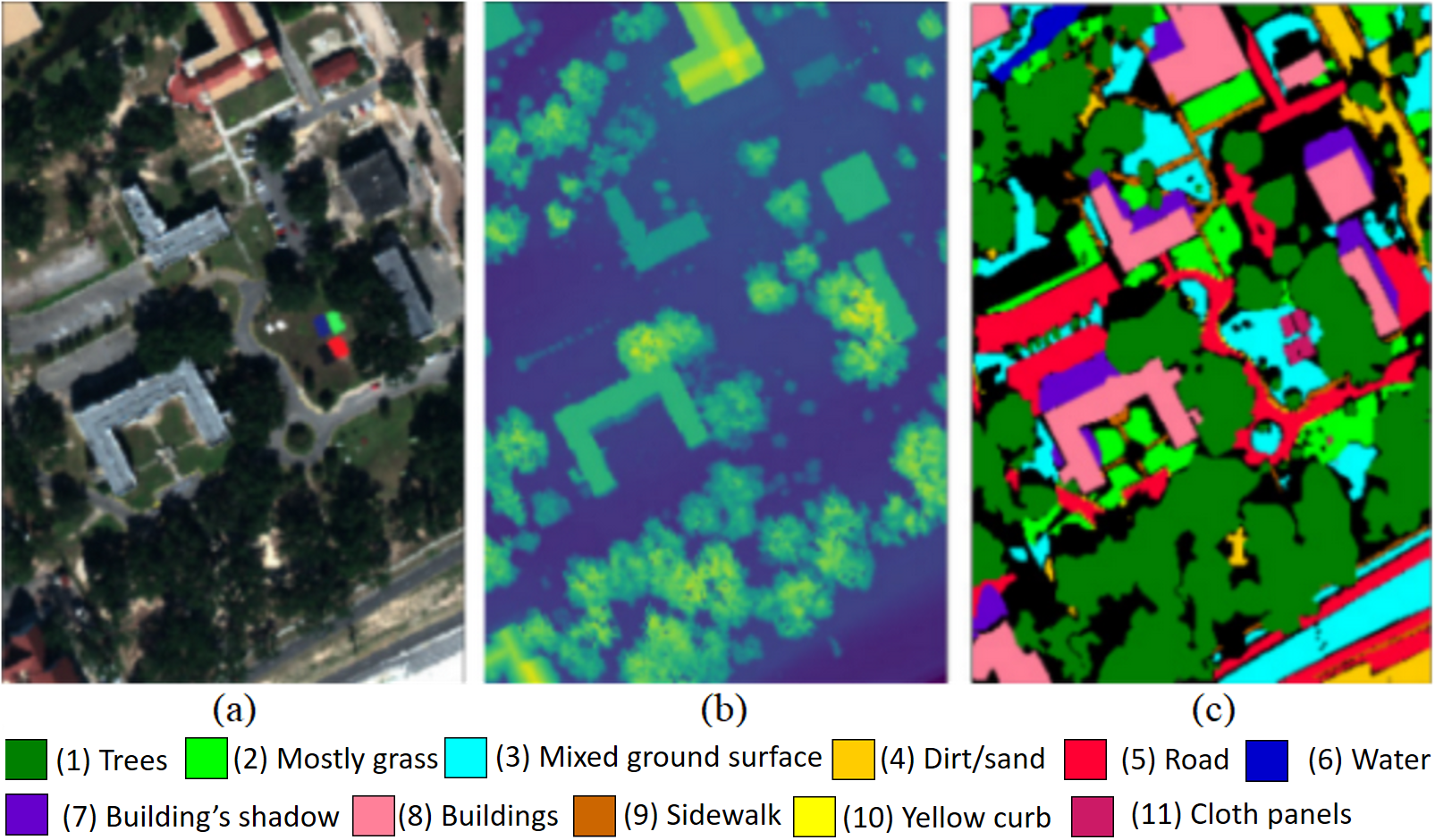}
    \caption{Dataset. (a) RGB FCC, (b) LiDAR, (c) Ground Truth}
    \label{fig:muufl}
\end{figure}

\subsection{MUUFL Gulfport dataset}

This is an HSI + LiDAR dataset acquired over the campus of University of Southern Mississippi Gulf Park, Long Beach Mississippi. The HSI data is composed of a total of 64 bands, while the LiDAR modality has one raster each corresponding to elevation and intensity. All the images are coregistered to the size of 325$\times$220. A total of 53687 ground truth pixels are distributed amongst 11 classes (Fig. \ref{fig:muufl}). For training purposes, 100 pixels per class are selected leaving the rest of 52587 pixels for testing \cite{mohla2020fusatnet}. The results in Table \ref{tab:MU_perf} show major improvements in classes 2, 6 and 10 and marginal improvements in classes 1, 2 and 5.

\subsection{Ablation study}

Ablation studies were performed to analyze the variation on change of the stack number and as well as the embedding dimension. The embedding dimension was fixed to 128 while variation of stack number was performed. The stack number $N_x$ was kept 4 while varying the embedding dimension. It can be seen that increasing embedding dimension does increase the accuracy to a specific number but beyond the threshold number, its increase does not contribute significant improvement in the accuracy (Table \ref{tab:abl2}). However, we can see an increase in accuracy as the stack number increases (Table \ref{tab:abl1}). This shows that self-attention along with cross-modal interaction helps in identifying features for better classification.

\section{CONCLUSIONS AND FUTURE WORKS}
\label{sec:conclusion}

In this paper we propose a novel cross-modal interaction based fusion model which tries to leverage the two complementary modalities for achieving higher accuracies. The results show that by increasing the interaction, the model has an increased overall accuracy as it can identify features better due to the cross-modal attention mechanism and infer more information via self-attention. The concatenation aides in information retention in an efficient encapsulation form. We would like to further extend this architecture as a scope to introduce more than two modalities. Due to modular nature, and cross-modal attention being inherent, it can open up new avenues in solving other cross-domain tasks more efficiently.

\section{ACKNOWLEDGEMENTS}
\label{sec:AC}

The authors would like to thank ISRO for the grant RD/0120-ISROC00-005 that contributed to this research. 

\bibliographystyle{IEEEbib}
\bibliography{strings,refs}

\end{document}